\documentclass[10pt,twocolumn,letterpaper]{article}

\usepackage{iccv}
\usepackage{times}
\usepackage{epsfig}
\usepackage{graphicx}
\usepackage{amsmath}
\usepackage{amssymb}

\usepackage[british, english, american]{babel}
\usepackage{pifont}
\usepackage{color, colortbl}
\usepackage{booktabs,tabulary,multirow,overpic,xcolor}
\usepackage[caption=false]{subfig}
\usepackage{amsthm}

\usepackage[pagebackref=true, breaklinks=true, colorlinks,
linkcolor=linkcolor, bookmarks=false]{hyperref}

\iccvfinalcopy 

\def\iccvPaperID{****} 

\ificcvfinal\fi

\newlength\savewidth\newcommand\shline{\noalign{\global\savewidth\arrayrulewidth
  \global\arrayrulewidth 1pt}\hline\noalign{\global\arrayrulewidth\savewidth}}
\newcommand{\tablestyle}[2]{\setlength{\tabcolsep}{#1}\renewcommand{\arraystretch}{#2}\centering\footnotesize}

\definecolor{LightCyan}{rgb}{0.88,1,1}

\begin{document}
\newcommand{\cmark}{\ding{51}}%
\newcommand{\xmark}{\ding{55}}%
\definecolor{linkcolor}{HTML}{ED1C24}
\definecolor{Gray}{gray}{0.9}
\definecolor{defaultchoice}{gray}{.95}
\definecolor{deltacolor}{gray}{.95}
\newcommand{\deltacolor}[1]{\cellcolor{deltacolor}{#1}}
\newcommand{\defaultchoice}[1]{\cellcolor{defaultchoice}{#1}}
\definecolor{impro}{gray}{.95}
\newcommand{\impro}[1]{\cellcolor{impro}{#1}}

\title{Enhancing Your Trained DETRs with Box Refinement}

\author{Yiqun Chen$^{1,2}$ \quad Qiang Chen$^{3}$ \quad Peize Sun$^{4}$ \quad Shoufa Chen$^{4}$ \quad Jingdong Wang$^{3}$ \quad Jian Cheng$^{1}$\\
$^{1}$Institute of Automation, Chinese Academy of Sciences\\
$^{2}$School of Artificial Intelligence, University of Chinese Academy of Sciences\\
$^{3}$Baidu\\
$^{4}$The University of Hong Kong
}

\maketitle
\ificcvfinal\thispagestyle{empty}\fi

\begin{abstract}
   We present a conceptually simple, efficient, and general framework for localization problems in DETR-like models. We add plugins to well-trained models instead of inefficiently designing new models and training them from scratch. The method, called RefineBox, refines the outputs of DETR-like detectors by lightweight refinement networks. RefineBox is easy to implement and train as it only leverages the features and predicted boxes from the well-trained detection models. Our method is also efficient as we freeze the trained detectors during training. In addition, we can easily generalize RefineBox to various trained detection models without any modification. We conduct experiments on COCO and LVIS $1.0$. Experimental results indicate the effectiveness of our RefineBox for DETR and its representative variants (Figure~\ref{fig:improvements}). For example, the performance gains for DETR, Conditinal-DETR, DAB-DETR, and DN-DETR are 2.4 AP, 2.5 AP, 1.9 AP, and 1.6 AP, respectively. We hope our work will bring the attention of the detection community to the localization bottleneck of current DETR-like models and highlight the potential of the RefineBox framework. Code and models will be publicly available at: \href{https://github.com/YiqunChen1999/RefineBox}{https://github.com/YiqunChen1999/RefineBox}.
\end{abstract}

\section{Introduction}
\label{sec:intro}

\begin{figure}[t]
    \begin{center}
       \includegraphics[trim={0.5cm 0.2cm 0.8cm 0.7cm}, clip, width=1.0\linewidth]{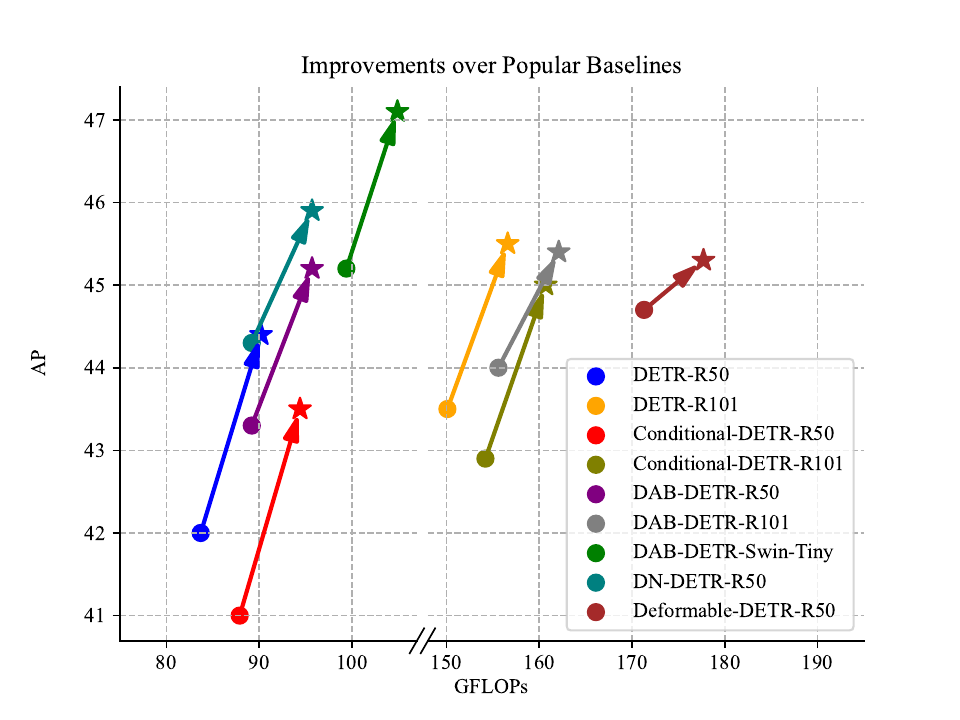}
    \end{center}
    \vspace{-1.2em}
   \caption{Our RefineBox framework gives non-trivial improvements over popular detectors on COCO~\cite{COCO-ECCV-2014}. The newly added refinement network only introduces 0.4 M parameters for Swin-Tiny-based detectors and 0.5 M additional parameters for ResNet-50/101-based models. Best viewed in color.}
\label{fig:improvements}
\end{figure}

Transformer~\cite{Transformer-NeurIPS-2017} has become an alternative to convolutional neural networks in object detection. DETR~\cite{DETR-ECCV-2020} introduces the Transformer into object detection and forms a new detection fashion. Recently, some DETR-like models became state-of-the-art on the COCO benchmark~\cite{COCO-ECCV-2014}, \eg, Co-DETR~\cite{Co-DETR-arXiv-2022}, Group-DETR-V2~\cite{Group-DETR-V2-arXiv-2022}, DETA~\cite{DETA-arXiv-2022}, DINO~\cite{DINO-ICLR-2022}, etc. Despite their success, the detection community might not be clear about the bottleneck of the DETR-like models and the potential development direction. 

The DETR-like detectors produce a set of predictions and assign each ground truth a prediction result during training. The positive samples, \ie, that match with ground truth, contribute significantly to the model training. \textit{We are curious about how to improve the prediction results of the DETR-like models by correcting the positive samples during training.} We start by exploring the performance upper bound of DETR-like models with perfect localization or classification ability in a simple way. Specifically, we calculate the performance gains brought by eliminating classification and localization errors of the positive samples. We observe that the gain of improving localization ability is substantial while improving classification capability is much lower, \ie, about 25 AP \vs 3-5 AP  (Figure~\ref{fig:motivation}, discussed later in Section~\ref{sec:motivation}). This phenomenon suggests that the localization capability is the bottleneck restricting the performance of current DETR-like models rather than the classification ability. 
{Based on the above finding, we focus on improving the localization ability of DETR-like models.}

\begin{figure}[t]
    \begin{center}
       \includegraphics[trim={8.4cm 9.0cm 8.4cm 9.0cm}, clip, width=1.0\linewidth]{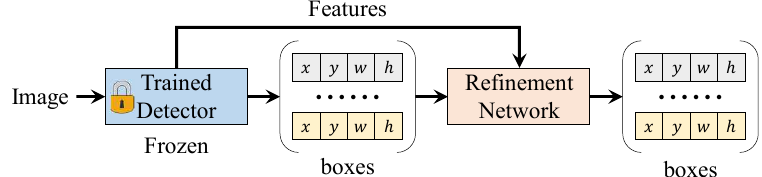}
    \end{center}
    \vspace{-1.0em}
   \caption{We build our RefineBox on top of trained and frozen DETR-like detection models. The refinement is responsible for refining the predicted boxes of the detectors. We only fit the parameters of the refinement network during training.}
\label{fig:framework-overview}
\end{figure}

\begin{figure}[!t]
    \begin{center}
       \includegraphics[trim={0.3cm 0.0cm 0.0cm 0.0cm}, clip, width=1.0\linewidth]{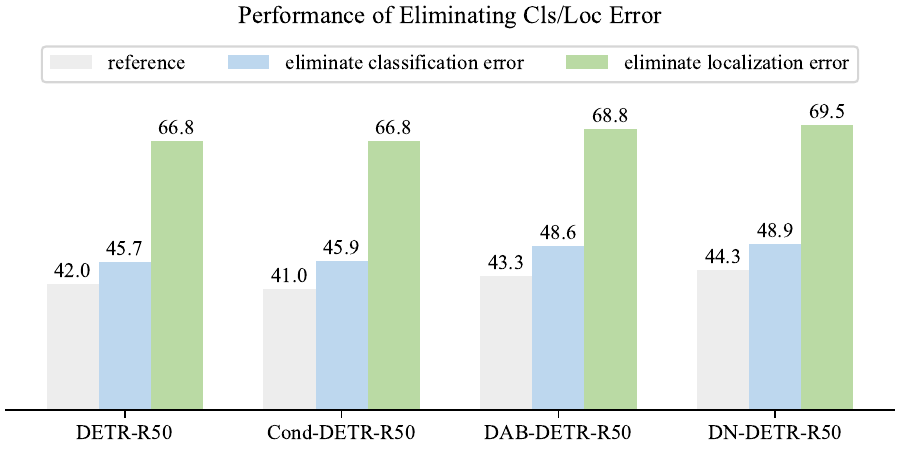}
    \end{center}
    \vspace{-1.0em}
   \caption{We obtain the ideal performance (AP) when eliminating the classification errors (blue bars) and localization errors (green bars). Eliminating the classification leads to perfect classification ability but improves the model less. Improving localization can bring impressive gains. }
\label{fig:motivation}
\end{figure}

We present a novel framework RefineBox to solve the localization problem in this paper. 
Our method, called RefineBox, adopts the two-stage detection framework by applying a lightweight box refinement network to the well-trained object detectors, as illustrated in Figure~\ref{fig:framework-overview}. The object detectors are responsible for producing the detection results, including the classification logits and the bounding boxes. 
{The extracted features and predicted bounding boxes from the trained detector serve as inputs to the refinement network, which is responsible for enhancing the boxes to mitigate localization errors.}

{RefineBox explores a new fashion for improving the localization quality of the trained detectors, which is different from previous works on two-stage detection that mainly design and train new detectors. Our refinement network acts like a plug-and-play plugin and brings clear benefits. It is easy to implement and train as we only need to fit the newly added parameters by freezing the trained detectors. Furthermore, it is highly efficient and cost-effective, facilitating speedy experimentation and the implementation of enhancements to large models with limited hardware. Moreover, RefineBox can be generalized to various detection models without any modification.}

{We validate the effectiveness of}
our method on DETR~\cite{DETR-ECCV-2020} and its representative variants, including Conditional-DETR-R50~\cite{Conditional-DETR-ICCV-2021}, DAB-DETR-R50~\cite{DAB-DETR-arXiv-2022}, and DN-DETR-R50~\cite{DN-DETR-CVPR-2022}. Without bells and whistles, our RefineBox brings non-trivial improvements to them, as illustrated in Figure~\ref{fig:improvements}. We also show that our RefineBox can easily combine with techniques that aim to speed up model convergence, \eg, Group-DETR~\cite{Group-DETR-arXiv-2022}. Our example gives a gain of 2.7 AP on Group-Conditional-DETR-R50 on the COCO dataset~\cite{COCO-ECCV-2014}.

\section{Related Works}
\noindent \textbf{Two-Stage Detection.} The two-stage detection architecture~\cite{Grid-R-CNN-CVPR-2019, Dynamic-R-CNN-ECCV-2020, TridentNet-iccv-2019, Libra-R-CNN-CVPR-2019} has become predominant several years ago due to the success of the models in the R-CNN series~\cite{RCNN-CVPR-2014, FastRCNN-ICCV-2015, FasterRCNN-NeurIPS-2015}. A typical two-stage detection model splits the detection pipeline into two stages: (1) produces a set of region proposals that might contain an object; (2) adjust the proposed bounding boxes and predict the categories. Our RefineBox takes the idea of two-stage detection but makes some changes: RefineBox builds on top of well-trained detection models and freezes them. 

\noindent \textbf{Error Analysis for Object Detection.} The community measures the detection performance by AP (Averaged Precision) and mAP (mean Average Precision). To further understand why we consider a detection a failure, Hoiem~\etal~\cite{hoiem2012diagnosing} introduced a categorization for false positive detections. 
The COCO~\cite{COCO-ECCV-2014} analysis tool further computes the numbers of several error types progressively. However, the analysis results rely on the computing order of the error types~\cite{TIDE-ECCV-2020}. TIDE~\cite{TIDE-ECCV-2020} then solves this issue by avoiding computing errors progressively. DCR~\cite{DCR-ECCV-2018, DCRv2-arXiv-2018} recognizes that the classification capability is responsible for the false positives of the Faster RCNN~\cite{FasterRCNN-NeurIPS-2015}. It combines the Faster RCNN and a classification refinement network to mitigate this issue. Some works~\cite{rahman2019did, ramanagopal2018failing, yang2021introspective, miller-cvpr-2022} focus on false negatives and solve this issue by introducing some prior. Miller~\etal~\cite{miller2022s} split the detection pipeline into several stages to locate the stage when a model fails to detect an object. 

\noindent \textbf{Freezing Detector's Parameters.} Recently, there are some works~\cite{FrozenBackbone-CVPR-2022, GiantPretrainedImageModels-arXiv-2022, Large-UniDet-arXiv-2022, F-VLM-arXiv-2022} on exploiting the frozen components on detection. Vasconcelos~\etal~\cite{FrozenBackbone-CVPR-2022} pay attention to freezing the backbone. They observed that with the powerful detection components, detection models trained with frozen backbones perform better than that with a non-frozen strategy. Lin~\etal~\cite{GiantPretrainedImageModels-arXiv-2022} further answer how to better apply this frozen setting. An open-vocabulary detection method, F-VLM~\cite{F-VLM-arXiv-2022}, freezes the vision and language models and only fits the detection head during training. Recently, Large-UniDet~\cite{Large-UniDet-arXiv-2022} freezes the backbone to handle the million-scale multi-domain universal object detection problem. Although RefineBox shares the similar idea of freezing trained models, it has several differences: (1) They aim to reuse the image classification features properly. In contrast, we explore a new fashion for improving localization quality. (2) They freeze the classification model while RefineBox freezes the whole detector. 

\begin{figure*}[!t]
\begin{center}
\includegraphics[trim={4.5cm 7.0cm 4.5cm 7.0cm}, clip, width=1.0\textwidth]{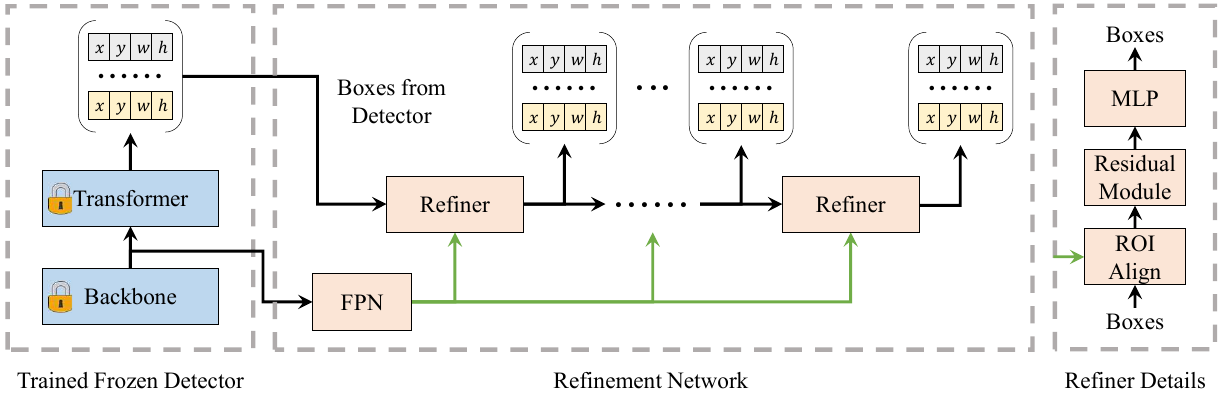}
\end{center}
\vspace{-1.5em}
   \caption{We show the details of our model. The refinement network leverages the feature pyramid produced from the backbone to refine the predicted boxes of the trained detector.}
\label{fig:overview}
\end{figure*}

\section{Motivation}
\label{sec:motivation}

The recent DETR-like models produce a set of predictions and adopt the one-to-one assignment strategy to find positive samples for training. We are eager to know how to improve the performance of DETR-like models by reducing the errors of positive samples during training. As the detection task consists of two sub-tasks: classification and localization, we try to answer this question by investigating the classification and localization errors of the positive samples. 

We start by finding predictions of an object detector that match the ground truth via Hungarian Matching~\cite{Hungarian-1955}. To calculate the performance with perfect localization, we replace the predicted bounding boxes with ground truth to eliminate the localization errors. Similarly, replacing predicted classification results with ground truth labels enables us to simulate models with perfect classification ability and calculate their ideal performance.

We visualize the performance of various DETR-like models~\cite{DETR-ECCV-2020, Conditional-DETR-ICCV-2021, DAB-DETR-arXiv-2022, DN-DETR-CVPR-2022} before and after replacing the predictions with the ground truth. We have two observations from Figure~\ref{fig:motivation}. \textit{First, eliminating the localization errors leads to impressive improvements in the detectors} (green bars \vs gray bars). We observe that the gains of most models are about 25 AP, \eg, 24.8 AP for DETR-R50~\cite{DETR-ECCV-2020}, 25.5 AP for DAB-DETR-50~\cite{DAB-DETR-arXiv-2022}, and 25.2 AP for DN-DETR-R50~\cite{DN-DETR-CVPR-2022}, etc. This indicates that there is significant potential to improve localization performance.

\textit{Second, eliminating classification errors leads to far smaller gains} (blue bars \vs gray bars), \eg, 3.7 AP for DETR-R50, 5.3 AP for DAB-DETR-R50, and 5.6 AP for DN-DETR-R50. This suggests that models with perfect classification ability can only provide limited performance gains for current models. The remaining errors are due to localization. Based on the above observations, we suppose localization is the bottleneck for current DETR-like models: The localization errors limit the possible benefits of improving the classification ability. Even models with ideal classification performance can only bring limited gains.

The above findings motivate our work to tackle the challenges of poor localization ability. We propose RefineBox, a novel framework for refining the predicted bounding boxes of existing detection models. We present a lightweight example in the following, which only introduces around 0.5 M parameters but produces non-trivial improvements, introduced next.

\section{Methodology}
\label{sec:methodology}

In this section, we first give more details of the RefineBox framework and the example in Section~\ref{subsec:method-details}. Then we present discussions on the relation to other methods in Section~\ref{subsec:method-discussion}.

\subsection{Method Details}
\label{subsec:method-details}

The proposed RefineBox (Figure~\ref{fig:framework-overview}) incorporates two components: (1) an object detector, and (2) a refinement network for refining the predicted bounding boxes. In our experiments, the refinement network (Figure~\ref{fig:overview}) mainly performs two steps. Firstly, by leveraging the feature pyramid network (FPN)~\cite{FPN-CVPR-2017}, the refinement network extracts the feature pyramid from the backbone of the trained detection model. Secondly, it efficiently leverages the multi-scale features via a sequence of Refiner modules to refine the boxes predicted by the detector. During training, we freeze the parameters of the trained object detector and only update the weights of the FPN and Refiner modules.

\noindent \textbf{Extracting the Multi-scale Features.} The feature pyramid of the detector backbone is utilized as the input to the FPN, which subsequently reduces the channels to a specific numerical value denoted as $C$ and is regarded as the model dimension. Specifically, for the ResNet-style~\cite{ResNet-2016} backbone, the inputs of FPN are $res_2$, $res_3$, $res_4$, and $res_5$. As for the Swin-style~\cite{Swin-Transformer-ICCV-2021}, $p_0, p_1, p_2$, and $p_3$ serve as the inputs to the FPN. Supplementary material provides ablation studies on the model dimension.

\noindent \textbf{Refining the Boxes.} Once we obtain the multi-scale features, we apply a series of Refiner modules to refine the predicted boxes, as depicted in Figure~\ref{fig:overview}. The Refiner module takes the boxes and the feature pyramid as inputs. Similar to Deformable DETR~\cite{Deformable-DETR-arXiv-2020}, we correct the boxes by predicting the deltas to the ground truth. Each Refiner module comprises an ROI Align layer~\cite{Mask-R-CNN-ICCV-2017}, a Residual Module~\cite{ResNet-2016}, and a multi-layer perceptron (MLP). Our Residual Module consists of several Bottleneck Blocks~\cite{ResNet-2016}. Each block stacks a $1 \times 1$ conv layer, a $3 \times 3$ conv layer, and a $1 \times 1$ conv layer. The bottleneck channels correspond to the input channels of the $3 \times 3$ conv layer. We find that sharing the weights of the Refiner modules does not harm the performance (see Section~\ref{subsec:ablation}), so their parameters are shared unless specified. Ablation on the number of Refiner modules, the bottleneck channels, and the number of Bottleneck Blocks are available in the supplementary material.

\noindent \textbf{Loss Functions.} Our RefineBox framework only refines the predicted bounding boxes and remains the classification results unchanged. Therefore, we only consider the regression loss in this framework. Specifically, we apply the GIoU loss~\cite{GIoU-CVPR-2019} and the L$_1$ loss to the outputs of each Refiner module. Following the common practice~\cite{DETR-ECCV-2020}, the weights of GIoU loss and L$_1$ loss remains unchanged. We sum the regression losses over all Refiner modules.

\noindent \textbf{Training and Inference.} During training, we select the predictions that match the ground truth and refine the corresponding bounding boxes. Other predictions that are not assigned ground truth are ignored. For inference, we select top $K$ predictions according to the classification scores for refinement, as it's impossible to access the ground truth. The value of $K$ is determined as the highest number of objects in an image in the dataset, \eg, 100 in COCO~\cite{COCO-ECCV-2014}. 

\noindent \textbf{Implementation Details.} Unless specified, we set the output channels $C$ of FPN~\cite{FPN-CVPR-2017} as 64 and the number of Bottleneck Blocks as 3. The output size of ROI Align layer is (7, 7). The default value of channels of $3 \times 3$ conv layers in Bottleneck blocks is 64. We refine the top 100 predictions during inference. The number of Refiner modules is 3.

\subsection{Discussion: Relation to Other Formulation.}
\label{subsec:method-discussion}

\noindent \textbf{Relation to Two-Stage/Cascade Detection. } In principle our RefineBox is a two-stage detection framework~\cite{RCNN-CVPR-2014, FastRCNN-ICCV-2015, FasterRCNN-NeurIPS-2015}. In the first stage, the detector produces initial predictions. In this stage, the detection model serves as a region proposal network. In the second stage, a refinement network takes the proposals as input and outputs the refined results. Though our RefineBox takes the two-stage detection pipeline, there are some differences from the typical two-stage detectors. In the first stage, the detector produces detection results instead of rejecting proposals that are unlikely to have an object. In addition, we build RefineBox on top of well-trained detectors and freeze them, and only train the newly added plug-and-play refinement network. 

Cascade detection~\cite{Cascade-R-CNN-CVPR-2018} is an extension of the two-stage detection. Several works have adopted a similar idea, \eg, IoUNet~\cite{IoU-Net-CVPR-2018}, Multi-Region CNN~\cite{Multi-Region-CNN-ICCV-2015}, ALFNet~\cite{ALFNet-ECCV-2018}, and BPN~\cite{BPN-Neurocomputing-2020}. The cascade detection architecture consists of a sequence of detection heads and is trained with increasing IoU thresholds. Thanks to the simplicity of our RefineBox, we can easily generalize it to a cascade refinement framework by stacking several refinement networks. It is worth noting that we still only need to fit the refinement network.

\noindent \textbf{Relation to Frozen Feature Extractors.} There are several works for freezing feature extractors~\cite{FrozenBackbone-CVPR-2022, F-VLM-arXiv-2022}, \eg, image and text feature extractors. One representative work is ~\cite{FrozenBackbone-CVPR-2022}. The method concentrates on freezing the backbone of object detectors. Although our approach also employs a similar idea of freezing model components, {the motivation behind our method is quite different}. Their methods aim to transfer the well-trained image/text feature extractor to object detection and utilize features produced by pre-trained models. In contrast, we aim to improve the localization ability of trained detection models efficiently. 

\begin{table*}[t]
\tablestyle{3.5pt}{1.1}
\begin{tabular}{l|l|rrrrrr|rrcrrr}
 Model & Backbone & AP & AP$_{50}$ & AP$_{75}$ & AP$_{s}$ & AP$_{m}$ & AP$_{l}$ & AR$_{1}$ & AR$_{10}$ & AR$_{100}$ & AR$_{s}$ & AR$_{m}$ & AR$_{l}$\\
\shline
DETR & ResNet-50 & 42.0 & 62.3 & 44.3 & 20.7 & 45.9 & 61.1 & 33.4 & 53.3 & 57.5 & 31.8 & 63.0 & 80.7 \\
+  RefineBox (Ours) & ResNet-50 & \textbf{44.4} & \textbf{63.2} & \textbf{47.6} & \textbf{24.7} & \textbf{47.6} & \textbf{61.5} & \textbf{34.8} & \textbf{56.4} & \textbf{61.2} & \textbf{38.4} & \textbf{65.7} & \textbf{81.1} \\
$\Delta$       && \deltacolor{+2.4} & \deltacolor{+0.9} & \deltacolor{+3.3} & \deltacolor{+4.0} & \deltacolor{+1.7} & \deltacolor{+0.4} & \deltacolor{+1.4} & \deltacolor{+3.1} & \deltacolor{+3.7} & \deltacolor{+6.6} & \deltacolor{+2.7} & \deltacolor{+0.4}  \\
\hline 
DETR & ResNet-101 & 43.5 & 63.8 & 46.5 & 22.0 & 48.0 & 61.8 & 34.5 & 54.9 & 59.0 & 33.8 & 64.5 & 81.4 \\
+  RefineBox (Ours) & ResNet-101 & \textbf{45.5} & \textbf{64.4} & \textbf{49.0} & \textbf{25.3} & \textbf{49.4} & \textbf{62.3} & \textbf{35.5} & \textbf{57.5} & \textbf{62.1} & \textbf{39.3} & \textbf{66.7} & \textbf{81.8} \\
$\Delta$        && \deltacolor{+2.0} & \deltacolor{+0.6} & \deltacolor{+2.5} & \deltacolor{+3.3} & \deltacolor{+1.4} & \deltacolor{+0.5} & \deltacolor{+1.0} & \deltacolor{+2.6} & \deltacolor{+3.1} & \deltacolor{+5.5} & \deltacolor{+2.2} & \deltacolor{+0.4} \\
\hline
Conditional-DETR & ResNet-50 & 41.0 & 62.1 & 43.4 & 20.6 & 44.5 & 59.5 & 33.7 & 55.4 & 61.0 & 35.7 & 67.0 & 84.4 \\
 +  RefineBox (Ours) & ResNet-50 & \textbf{43.5} & \textbf{63.1} & \textbf{46.8} & \textbf{24.4} & \textbf{46.5} & \textbf{60.1} & \textbf{35.2} & \textbf{58.8} & \textbf{65.1} & \textbf{43.2} & \textbf{70.0} & \textbf{85.4} \\
$\Delta$               && \deltacolor{+2.5} & \deltacolor{+1.0} & \deltacolor{+3.4} & \deltacolor{+3.8} & \deltacolor{+2.0} & \deltacolor{+0.6} & \deltacolor{+1.5} & \deltacolor{+3.4} & \deltacolor{+4.1} & \deltacolor{+7.5} & \deltacolor{+3.0} & \deltacolor{+1.0} \\
\hline
Conditional-DETR & ResNet-101  & 42.9 & 63.9 & 45.9 & 22.1 & 46.9 & 60.9 & 34.7 & 56.7 & 62.1 & 38.2 & 68.1 & 85.2 \\
+  RefineBox (Ours) & ResNet-101 & \textbf{45.0} & \textbf{64.7} & \textbf{48.3} & \textbf{25.5} & \textbf{48.3} & \textbf{61.4} & \textbf{35.8} & \textbf{59.6} & \textbf{65.6} & \textbf{44.5} & \textbf{70.5} & \textbf{85.7} \\
$\Delta$               && \deltacolor{+2.1} & \deltacolor{+0.8} & \deltacolor{+2.4} & \deltacolor{+3.4} & \deltacolor{+1.4} & \deltacolor{+0.5} & \deltacolor{+1.1} & \deltacolor{+2.9} & \deltacolor{+3.5} & \deltacolor{+6.3} & \deltacolor{+2.4} & \deltacolor{+0.5} \\
\hline
DAB-DETR & ResNet-50           & 43.3 & 63.9 & 45.9 & 23.4 & 47.1 & 62.1 & 34.9 & 57.5 & 62.9 & 39.2 & 68.9 & 85.6\\
+  RefineBox (Ours) & ResNet-50 & \textbf{45.2} & \textbf{64.5} & \textbf{48.6} & \textbf{26.6} & \textbf{48.2} & \textbf{62.4} & \textbf{36.0} & \textbf{60.0} & \textbf{65.9} & \textbf{44.8} & \textbf{70.7} & \textbf{85.8}\\
$\Delta$               && \deltacolor{+1.9} & \deltacolor{+0.6} & \deltacolor{+2.7} & \deltacolor{+3.2} & \deltacolor{+1.1} & \deltacolor{+0.3} & \deltacolor{+1.1} & \deltacolor{+2.5} & \deltacolor{+3.0} & \deltacolor{+5.6} & \deltacolor{+1.8} & \deltacolor{+0.2} \\
\hline
DAB-DETR & ResNet-101          & 44.0 & 62.9 & 47.6 & 23.8 & 48.4 & 61.9 & 35.3 & 59.0 & 65.7 & 42.3 & 72.1 & 86.5 \\
+  RefineBox (Ours) & ResNet-101 & \textbf{45.4} & \textbf{63.2} & \textbf{49.6} & \textbf{26.2} & \textbf{49.3} & \textbf{62.1} & \textbf{36.1} & \textbf{61.0} & \textbf{68.4} & \textbf{47.6} & \textbf{73.5} & \textbf{86.8} \\
$\Delta$               && \deltacolor{+1.4} & \deltacolor{+0.3} & \deltacolor{+2.0} & \deltacolor{+2.4} & \deltacolor{+0.9} & \deltacolor{+0.3} & \deltacolor{+0.8} & \deltacolor{+2.0} & \deltacolor{+2.7} & \deltacolor{+5.3} & \deltacolor{+1.4} & \deltacolor{+0.3}  \\
\hline
DAB-DETR & Swin-Tiny     & 45.2 & 66.8 & 47.8 & 24.2 & 49.0 & 64.8 & 35.8 & 58.4 & 63.5 & 40.2 & 69.6 & 86.5 \\
+  RefineBox (Ours) & Swin-Tiny & \textbf{47.1} & \textbf{67.5} & \textbf{50.5} & \textbf{27.6} & \textbf{50.1} & \textbf{65.1} & \textbf{36.7} & \textbf{60.9} & \textbf{66.5} & \textbf{46.5} & \textbf{70.9} & \textbf{86.6} \\
$\Delta$               && \deltacolor{+1.9} & \deltacolor{+0.7} & \deltacolor{+2.7} & \deltacolor{+3.4} & \deltacolor{+1.1} & \deltacolor{+0.3} & \deltacolor{+ 0}.9 & \deltacolor{+2.5} & \deltacolor{+3.0} & \deltacolor{+6.3} & \deltacolor{+1.3} & \deltacolor{+0.1}\\
\hline
DN-DETR & ResNet-50            & 44.3 & 64.8 & 47.3 & 23.8 & 48.2 & 63.2  & 35.4 & 58.3 & 63.4 & 40.0 & 69.5 & 85.9  \\
+  RefineBox (Ours) & ResNet-50 & \textbf{45.9} & \textbf{65.3} & \textbf{49.4} & \textbf{26.2} & \textbf{49.2} & \textbf{63.5} & \textbf{36.2} & \textbf{60.3} & \textbf{66.0} & \textbf{44.9} & \textbf{70.8} & \textbf{86.0} \\
$\Delta$               && \deltacolor{+1.6} & \deltacolor{+0.5} & \deltacolor{+2.1} & \deltacolor{+2.4} & \deltacolor{+1.0} & \deltacolor{+0.3} & \deltacolor{+0.8} & \deltacolor{+2.0} & \deltacolor{+2.6} & \deltacolor{+4.9} & \deltacolor{+1.3} & \deltacolor{+0.1}  \\
\hline
Deformable-DETR & ResNet-50    & 44.7 & 64.0 & 48.8 & 27.0 & 47.9 & 59.8 & 35.5 & 60.1 & 66.0 & 46.0 & 70.7 & 84.5 \\
+  RefineBox (Ours) & ResNet-50 & \textbf{45.3} & \textbf{64.1} & \textbf{49.5} & \textbf{27.8} & \textbf{48.3} & \textbf{59.9} & \textbf{35.9} & \textbf{61.0} & \textbf{67.2} & \textbf{48.0} & \textbf{71.5} & \textbf{84.7} \\
$\Delta$            &   & \deltacolor{+0.6} & \deltacolor{+0.1} & \deltacolor{+0.7} & \deltacolor{+0.8} & \deltacolor{+0.4} & \deltacolor{+0.1} & \deltacolor{+0.4} & \deltacolor{+0.9} & \deltacolor{+1.2} & \deltacolor{+2.0} & \deltacolor{+0.8} & \deltacolor{+0.2}  \\
\end{tabular}\vspace{2mm}
\caption{Our RefineBox improves the AP and AR over DETR and its popular variants. The method mainly improves the AP, AP$_{75}$, and AP$_{s}$. We also observe significant increases in recalls, especially for AR$_{100}$ and AR$_{s}$.}
\label{tab:results-coco}\vspace{-3mm}
\end{table*}

\section{Experimental Results}
\label{sec:results}
In this section, we present the experimental results to validate the effectiveness of our RefineBox. 

\subsection{Settings}
\label{subsec:}

We conduct our experiments based on PyTorch~\cite{pytorch-NeurIPS-2019}, Detectron2~\cite{detectron2-2019} and detrex~\cite{detrex-2022}. Unless specified, we train our model for 12 epochs (90k iterations). We consider DETR~\cite{DETR-ECCV-2020} and its several popular variants as our baselines. We follow the training settings of baselines in detrex implementation to set the optimization algorithm and other hyperparameters. 

\noindent \textbf{Datasets.} We adopt the following two challenging benchmarks: 
(1) COCO~\cite{COCO-ECCV-2014}: The challenging COCO dataset contains about 118k images for training (train2017 split) and 5k images for validation (val2017). We mainly conduct experiments on this dataset. (2) LVIS~\cite{LVIS-CVPR-2019}: LVIS contains about 100k images for training and about 19.8k images for validation.

\subsection{Main Results}
We conduct experiments on the challenging COCO and LVIS $1.0$ datasets to demonstrate the effectiveness of our RefineBox.

\subsubsection{Results on COCO}
\noindent \textbf{Improvements Over Various Baselines.} We present the results on the COCO dataset in Table~\ref{tab:results-coco}, our  RefineBox significantly improves DETR~\cite{DETR-ECCV-2020} and its variants, \eg, Conditional-DETR~\cite{Conditional-DETR-ICCV-2021}, DAB-DETR~\cite{DAB-DETR-arXiv-2022}, and DN-DETR~\cite{DN-DETR-CVPR-2022}. Specifically, our method brings a 2.4 AP gain to DETR-R50 (from 42.0 AP to 44.4 AP). Besides, the  RefineBox gives 2.5 AP, 1.9 AP, and 1.6 AP  improvements to Conditional-DETR-R50, DAB-DETR-R50, and DN-DETR-R50, respectively. Remarkably, our approach only adjusts the bounding boxes and keeps the classification results unchanged. This phenomenon indicates our RefineBox is a promising framework.

We also note that our RefineBox considerably enhances recall rates. For example, the gains on AR$_{100}$ for DETR-R50 and Conditional-DETR-R50 are 3.7 and 4.1, respectively. The improvements on AR$_s$ are more impressive: 6.6 for DETR-R50 and 7.5 for Conditional-DETR-R50. 

\noindent \textbf{RefineBox and More Powerful Backbones.} To further demonstrate robustness and effectiveness, we explore the influence of a detector with a more powerful backbone, \eg, ResNet-101~\cite{ResNet-2016} and Swin-Tiny~\cite{Swin-Transformer-ICCV-2021}. Results presented in Table~\ref{tab:results-coco} show that our  RefineBox still works. With the ResNet-101 as the backbone of the object detector, our  RefineBox gives 2.0 AP, 2.1 AP, and 1.4 AP improvements on the DETR, Conditional-DETR, and DAB-DETR, respectively. Besides, we also observe 1.9 AP gains on DAB-DETR-Swin-Tiny. 

\begin{table*}[!t]
\tablestyle{3.5pt}{1.1}
\begin{tabular}{l|rrr|rrr|rrr|rrrr}
        Model & AP & AP$_{50}$ & AP$_{75}$ & AP$_{s}$ & AP$_{m}$ & AP$_{l}$ & AP$_{r}$ & AP$_{c}$ & AP$_{f}$ & AR & AR$_s$ & AR$_m$ & AR$_l$  \\
        \shline
        DAB-DETR-R50           & 19.9 & 31.0 & 20.8 & 12.1 & 28.1 & 35.8 & 8.0 & 19.1 & 26.0 & 30.1 & 17.4 & 40.4 & 56.3 \\
         +  RefineBox (Ours) & \textbf{21.8} & \textbf{31.5} & \textbf{22.8} & \textbf{14.6} & \textbf{29.8} & \textbf{36.6} & \textbf{8.3} & \textbf{20.7} & \textbf{29.0} & \textbf{32.8} & \textbf{20.8} & \textbf{42.9} & \textbf{57.3} \\
        $\Delta$               & \deltacolor{+1.9} & \deltacolor{+0.5} & \deltacolor{+2.0} & \deltacolor{+2.5} & \deltacolor{+1.7} & \deltacolor{+0.8} & \deltacolor{+0.3} & \deltacolor{+1.6} & \deltacolor{+3.0} & \deltacolor{+2.7} & \deltacolor{+3.4} & \deltacolor{+2.5} & \deltacolor{+1.0} \\
        \hline
        DAB-DETR-R50 + FedLoss   & 26.0 & 40.9 & 27.3 & 15.6 & 35.3 & 45.4 & 17.9 & 25.6 & 30.1 & 37.6 & 21.2 & 49.1 & 64.6 \\
         +  RefineBox (Ours) & \textbf{28.8} & \textbf{41.8} & \textbf{30.0} & \textbf{19.2} & \textbf{37.9} & \textbf{46.7} & \textbf{18.7} & \textbf{28.1} & \textbf{34.0} & \textbf{41.4} & \textbf{26.0} & \textbf{52.4} & \textbf{66.2} \\
        $\Delta$               & \deltacolor{+2.8} & \deltacolor{+0.9} & \deltacolor{+2.7} & \deltacolor{+3.6} & \deltacolor{+2.6} & \deltacolor{+1.3} & \deltacolor{+0.8} & \deltacolor{+2.5} & \deltacolor{+3.9} & \deltacolor{+3.8} & \deltacolor{+4.8} & \deltacolor{+3.3} & \deltacolor{+1.6} \\
\end{tabular}\vspace{2mm}
\caption{Our RefineBox significantly improves DAB-DETR-R50 on the LVIS $1.0$ dataset, especially for AP$_{75}$, AP$_s$, AP$_f$, and recalls. FedLoss is the federated loss.}
\label{tab:results-lvis-v1}\vspace{-3mm}
\end{table*}

\subsubsection{Results on LVIS 1.0}

\begin{table}[t]
\tablestyle{3.5pt}{1.1}
\begin{tabular}{l|rrr|rrrr}
        Model & AP$_{r}$ & AP$_{c}$ & AP$_{f}$ & AR & AR$_s$ & AR$_m$ & AR$_l$ \\
        \shline
        Reference  & 17.9     & 25.6     & 30.1   & 37.6 & 21.2 & 49.1 & 64.6\\
        \hline
        DAB-DETR-R50 & 50.8   & 43.8   & 36.0 & 50.5 & 30.9 & 63.7 & 79.8   \\
        +  RefineBox & \textbf{54.8} & \textbf{48.7} & \textbf{41.0} & \textbf{55.7} & \textbf{37.8} & \textbf{68.1} & \textbf{81.7} \\
        $\Delta$     & \deltacolor{+4.0}     & \deltacolor{+4.9}     & \deltacolor{+5.0}     & \deltacolor{+5.2}     & \deltacolor{+6.9}     & \deltacolor{+4.4}     & \deltacolor{+1.9}   \\
\end{tabular}\vspace{2mm}
\caption{The ideal performance of DAB-DETR-R50 (with federated loss) with perfect classification ability. Our method can effectively alleviate the localization problem in the detection model. For rare categories, the classification ability is also the bottleneck. In contrast, for the frequently occurred classes, the localization capability is the main problem.}
\label{tab:results-lvis-upper-bound}\vspace{-3mm}
\end{table}

We also conduct experiments on the LVIS 1.0 dataset~\cite{LVIS-CVPR-2019} to evaluate our method. Firstly, we train DAB-DETR-R50 from scratch on LVIS without performing any modification. All hyperparameters except the ones related to the number of categories are the same as those on the COCO dataset. We then freeze the DAB-DETR-R50 and apply our RefineBox. We also apply the federated loss~\cite{FederatedLoss-arXiv-2021} to the baseline to improve the detection results.

We have similar findings on LVIS as COCO from Table~\ref{tab:results-lvis-v1}: (1) our RefineBox improves the baseline by producing more accurate bounding boxes. For example, the AP increased from 26.0 to 28.8 AP (+ 2.8 AP); (2) the gains mainly come from the AP$_{75}$ and AP$_{s}$; (3) the improvements on the recall are significant. These phenomena are consistent with that on the COCO dataset and imply our method improves the localization quality. 

Interestingly, our RefineBox gives limited gains on rare classes (+ 0.8 AP) compared to the frequent categories (+ 3.9 AP). For better understanding, we eliminate the classification errors (refer to Section~\ref{sec:motivation} for more details) and present the ideal performance in Table~\ref{tab:results-lvis-upper-bound}. We observe that: (1) Our RefineBox increases the upper bound of AP$_r$, AP$_c$, and AP$_f$ by 4.0 AP, 4.9 AP, and 5.0 AP, respectively, which are comparable. (2) The upper bound of rare classes is much higher than the actual AP$_r$, no matter if we apply the  RefineBox. {These phenomena imply that the classification capability of the rare classes is one of the bottlenecks on LVIS: our method improves the localization quality, but detection fails due to the wrong classification results of DAB-DETR-R50. }

\begin{table*}[t]\vspace{-3mm}\hspace{0.5em}
\subfloat[Our RefineBox is orthogonal to Group DETR: the RefineBox brings significant gains on Group-Conditional-DETR-R50.\label{tab:results-group-detr}]{
\tablestyle{2.5pt}{1.05}\begin{tabular}{l|rrrrr}
  Model & AP & AP$_{75}$ & AP$_{s}$ & AP$_{m}$ & AP$_{l}$ \\
\shline
 Group-Cond-DETR & 37.6 & 39.2 & 17.4 & 40.8 & 55.5     \\
 +  RefineBox & \textbf{40.3} & \textbf{43.2} & \textbf{22.1} & \textbf{43.3} & \textbf{56.5}     \\
 $\Delta$   & \deltacolor{+2.7} & \deltacolor{+4.0} & \deltacolor{+4.7} & \deltacolor{+2.5} & \deltacolor{+1.0}    \\
\end{tabular}}\hspace{3mm}
\subfloat[The key reason for the performance gains is neither the extra parameters nor the additional FLOPs: more decoder layers do not improve the detector. We train all models for 12 epochs. \#dec represents the number of decoder layers. \label{tab:results-ablation-more-layers}]{
\tablestyle{4.8pt}{1.05}\begin{tabular}{l|r|rrrrr|rr}
 Model & \#dec & AP & AP$_{75}$ & AP$_{s}$ & AP$_{m}$ & AP$_{l}$ & Params & GFLOPs \\
\shline
DAB-DETR-R50+ & 12 & 35.1 & 36.5 & 16.1 & 38.1 & 52.8 & 54.7 M & 95.9\\
DAB-DETR-R50 & 6 & 35.9 & 37.4 & 17.3 & 38.9 & 53.6 & 43.7 M & 89.2\\
+ RefineBox & 6 & \textbf{39.2} & \textbf{41.8} & \textbf{22.6} & \textbf{41.7} & \textbf{54.9} & 44.2 M & 95.7\\
\end{tabular}}\vspace{-1mm}\\
\subfloat[The main reason for the performance gains is not the extra training time. 50+12: We first train the detector 50 epochs and then apply our RefineBox for 12 epochs.\label{tab:results-ablation-epochs-comparison}]{
\tablestyle{4pt}{1.05}\begin{tabular}{l|r|rrr}
 Model & Epochs & AP & AP$_{75}$ & AP$_{s}$ \\
\shline
Cond-DETR-R50 &  50 & 41.0 & 43.4 & 20.6  \\
Cond-DETR-R50 & 108   & 43.0 & 45.7 & 22.7 \\
RefineBox & 50+12 & \textbf{43.5} & \textbf{46.8} & \textbf{24.4} \\
\end{tabular}}\hspace{2mm}
\subfloat[Adding an additional classification branch does not help the final performance of our RefineBox. Both models report a result of 43.5 AP. We discuss this phenomenon in context. \label{tab:results-cls-branch}]{
\tablestyle{4pt}{1.05}\begin{tabular}{l|r|rrr}
 Model & cls & AP & AP$_{75}$ & AP$_{s}$ \\
\shline
 Cond-DETR-R50 & - & 41.0 & 43.4 & 20.6  \\
+  RefineBox & \xmark & \textbf{43.5} & {46.8} & {24.4} \\
+  RefineBox & \cmark &\textbf{ 43.5} & \textbf{46.9} & \textbf{24.5} \\
\end{tabular}}\hspace{2mm}
\subfloat[\textbf{T}raining the object \textbf{D}etector together does not improve the performance: freezing the detection model reports 45.2 AP, which is 0.3 AP higher than that of training detectors together.\label{tab:results-joint-training}]{
\tablestyle{4pt}{1.05}\begin{tabular}{l|r|rrr}
     Model & TD  & AP & AP$_{75}$ & AP$_{s}$ \\
    \shline
    DAB-DETR-R50 & -      & 43.3 & 45.9 & 23.4 \\
    + RefineBox  & \cmark & 44.9 & 48.1 & 25.7\\
    + RefineBox  & \xmark & \textbf{45.2} & \textbf{48.6} & \textbf{26.6} \\
\end{tabular}}\vspace{2mm}
\caption{Going deep with our RefineBox. We train on the train2017 split, and test on the val2017 subset.}
\label{tab:going-deeper}\vspace{-3mm}
\end{table*}

\subsection{Going Deeper}
We delve deeper into exploring an extension of our model and the rationale behind the effectiveness of our approach. We first present that RefineBox improves the ideal performance (\ie, error-free classification). We then demonstrate that our RefineBox can work with techniques designed to speed up model convergence. We also discuss several possible factors of the success of our model: (1) Whether the gains originate from the additional parameters or FLOPs; (2) Whether the gains come from the extra training time. Additionally, we consider: (1) Would an additional classification branch lead to further improvement? (2) would joint training with the detector aid in enhancing the model's performance? Finally, we hypothesize that the efficient utilization of multi-scale features may contribute to the success of the refinement network.

\begin{figure}[t]
\centering
\includegraphics[width=1.0\linewidth]{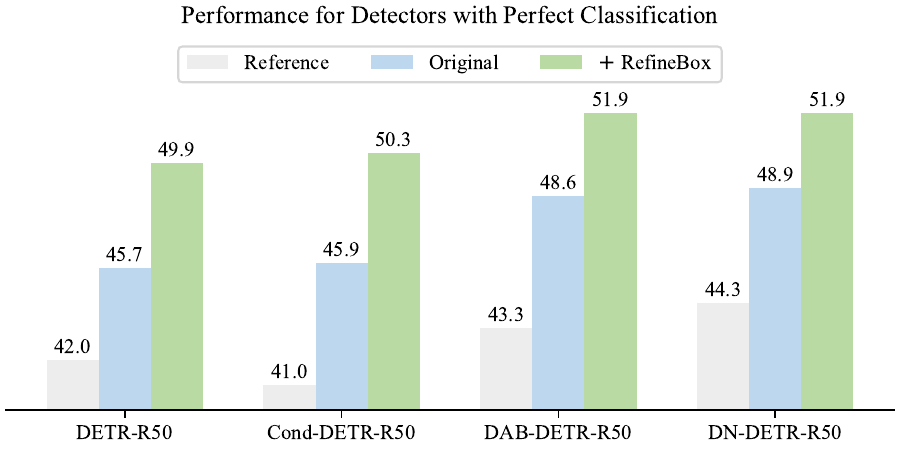}
\caption{RefineBox improves the ideal performance (AP) for DETR and its variants by reducing localization errors. \textit{Original}: eliminate classification errors but do not apply RefinBox. \textit{+RefineBox}: eliminate classification errors and apply RefineBox.}
\label{fig:upper-bound-gains}\vspace{-4mm}
\end{figure}

\noindent \textbf{RefineBox Increases the Ideal Performance.} Following how we locate the bottleneck of detection models (Section~\ref{sec:motivation}), we replace the labels with ground truth to investigate the ideal performance (the case without classification errors). Figure~\ref{fig:upper-bound-gains} suggests that our RefineBox effectively reduces the localization errors and improves the ideal performance of detection models. The results suggest that enhancing classification ability would likely result in further gains. We leave this exploration as future work. 

\noindent \textbf{Improvements on Group DETR}
Our method is orthogonal to the recently proposed Group DETR~\cite{Group-DETR-arXiv-2022} and can further boost the latter's performance. Specifically, we train Conditional-DETR-R50 with 11 groups for 12 epochs and then apply our RefineBox. As illustrated in Table~\ref{tab:results-group-detr}, our RefineBox improves the AP, AP$_{75}$, AP$_{s}$, and AP$_{m}$ of the baseline by 2.7 AP, 4.0 AP, 4.7 AP, and 2.5 AP, respectively.

\noindent \textbf{Do the gains come from the additional parameters or FLOPs?} We compare our RefineBox with DAB-DETR-R50~\cite{DAB-DETR-arXiv-2022} with 12 decoder layers (DAB-DETR-R50+) in Table~\ref{tab:results-ablation-more-layers}. The DAB-DETR-R50+ is designed to have comparable GFLOPs with our refinement network. We train all models for 12 epochs. The model with only six decoder layers gives 35.9 AP while adding another six decoder layers slightly degrades the performance, \ie, AP decreases to 35.1. In contrast, applying our RefineBox to DAB-DETR-R50 increases the score from 35.9 AP to 39.2 AP (+3.3 AP). \textit{In summary, the key reason for the performance gains is neither the extra parameters nor the additional FLOPs.}

Besides the performance advantage, our RefineBox also shows the efficiency of the parameters. Compared to our method's 0.5 M, DAB-DETR-R50+ introduces an additional 11 M parameters, 22 times of ours. Please note that all layers in DAB-DETR and RefineBox only update 12 epochs in all experiments, so the comparison is fair. Also, freezing DAB-DETR and adding six decoder layers is an example of our RefineBox framework. 

\noindent \textbf{Do the gains come from the extra training time?} We present the results of object detectors without our RefineBox but trained longer in Table~\ref{tab:results-ablation-epochs-comparison}. We observe that: (1) With 108 training epochs, Conditional-DETR-R50 reports 2.0 AP improvements over the baseline (50 epochs). (2) Our RefineBox under 12 epochs training leads to a gain of 2.5 AP, surpassing the performance of the Conditional-DETR-R50 with 108 training epochs. \textit{Given these observations, we conclude that extra training time is not the primary reason for the performance gain.}

\noindent \textbf{Would an additional classification branch help?} Our RefineBox only refines the predicted boxes. We are curious if an additional classification branch improves performance. The construction of the classification branch is similar to the localization branch. Specifically, we add the Refiner modules after the FPN, in parallel with the localization branch. We don't share the parameters of the localization branch with the classification branch {(see supplementary material for more details)}. 

Table~\ref{tab:results-cls-branch} shows that adding an additional classification branch does not help the final performance. We suspect that: (1) The capacity of the classification branch is much smaller. A prior work aiming to reduce the confidence score of false positives, \ie, DCR~\cite{DCR-ECCV-2018}, improves the Faster R-CNN~\cite{FasterRCNN-NeurIPS-2015} by adopting a ResNet-152~\cite{ResNet-2016} as the refinement network. In contrast, our design is lightweight. A more powerful classifier may further increase the performance. (2) It might be difficult for current DETR-like models to improve the overall performance by increasing the classification ability, as discussed in Section~\ref{sec:motivation}. 

\noindent \textbf{Would training the detector together help?} Our design of freezing the object detection models during training is efficient. We also explore whether training the detector concurrently would improve performance. Results in Table~\ref{tab:results-joint-training} suggest that the performance slightly degrades, \ie, from 45.2 AP to 44.9 AP. We suspect the reason is the inconsistency between the well-trained model and the randomly initialized refinement network. For our RefineBox framework, the refinement network only needs to fit the output features of the frozen detector. Therefore, it cannot influence the feature distribution of the detection model. In contrast, training the detector together makes the two affect each other. The detector may shift the distribution of the features under the effect of the refinement network. The refinement network should fit a new distribution once the shift occurs. The training instability may cause the optimization problem. Given the above findings, we suppose that freezing detectors in our RefineBox is effective and efficient. 

\noindent \textbf{Effective Utilization of Multi-scale Features.} We observe that our example mainly increases the localization quality of AP$_{s}$ and AP$_{75}$ (see Table~\ref{tab:results-coco} and Table~\ref{tab:results-lvis-v1}). We attribute this finding to the efficient utilization of the multi-scale features. The proposed model utilizes larger-size features, which might be difficult for the original detection models partially because the self-attention complexity is quadratic. We hypothesize leveraging multi-scale features is a core factor in the success of the refinement network. It might be one of the reasons why the refinement network in our experiments gives relatively low gains to Deformable DETR~\cite{Deformable-DETR-arXiv-2020}: Deformable DETR applies a relatively more complex Deformable Attention mechanism to leverage the feature pyramid. Further improvements can be expected with more sophisticated refinement network designs.

\noindent \textbf{Summary.} From the above discussion, we conclude that: (1) The extra training time, parameters, and FLOPs are not the main reasons for the success of our method. (2) Training together with an additional classification branch or well-trained detectors is inefficient and brings no benefits.

\begin{table}[t]
\tablestyle{3.5pt}{1.1}
\begin{tabular}{l|cc}
        Model                 & DAB-DETR-R50 & RefineBox \\
        \shline
        Training Time (mins)  & 558          & 324\\
\end{tabular}\vspace{2mm}
\caption{The training time of RefineBox with producing features and boxes online is 58\% of DAB-DETR.}
\label{tab:results-training-time}\vspace{-3mm}
\end{table}

\begin{table*}[t]\vspace{-3mm}
\subfloat[Our method is lightweight: RefineBox only introduces 0.5 M parameters and 6.5 GFLOPs for DAB-DETR-R50/101. \label{tab:results-overhead}]{
\tablestyle{4pt}{1.05}\begin{tabular}{l|rr}
 Model & Params & GFLOPs \\
\shline
DAB-DETR-R50       & 43.7 M & 89.2 \\
 + RefineBox    & 44.2 M & 95.7 \\
 \hline
DAB-DETR-R101      & 62.6 M & 155.6 \\
 + RefineBox    & 63.1 M & 162.1 \\
 \hline
DAB-DETR-SwinT & 47.4 M &  99.4 \\
 + RefineBox    & 47.8 M & 104.9 \\
\end{tabular}}\hspace{2mm}
\subfloat[Additional training epochs give limited gains. We suspect the model capacity is the bottleneck (the refinement network brings only 0.5 M parameters). \label{tab:results-ablation-epochs}]{
\tablestyle{4pt}{1.05}\begin{tabular}{l|r|rrr}
 Model & Epochs & AP & AP$_{75}$ & AP$_{s}$ \\
\shline
 Cond-DETR-R50 & 50  & 41.0 & 43.4 & 20.6  \\
+ RefineBox    & +12 & 43.5 & 46.8 & 24.4 \\
+ RefineBox    & +24 & 43.9 & 47.1 & 24.8 \\
+ RefineBox    & +36 & 44.0 & 47.3 & 25.0 \\
+ RefineBox    & +50 & \textbf{44.1} & \textbf{48.6} & \textbf{26.7} \\
 \multicolumn{5}{c}{~}\\ 
\end{tabular}}\hspace{2mm}
\subfloat[Model performance increases as $K$ gets larger until $K$ exceeds the maximum number of objects that can appear in an image.\label{tab:results-ablation-topk}]{
\tablestyle{4pt}{1.05}\begin{tabular}{l|r|rrr|r}
     Model       & $K$ & AP & AP$_{75}$ & AP$_{s}$ & FLOPs \\
    \shline
    DAB-DETR    & -   & 43.3 & 45.9 & 23.4 & 89.2 \\
    \hline
    + RefineBox & 300 & \textbf{45.2} & \textbf{48.6} & \textbf{26.6} & 99.8 \\
    + RefineBox & 100 & \textbf{45.2} & \textbf{48.6} & \textbf{26.6} & 95.7 \\
    + RefineBox &  30 & 45.0 & 48.3 & 26.3 & 94.3 \\
    + RefineBox &  10 & 44.5 & 47.6 & 25.4 & 93.9 \\
     \multicolumn{6}{c}{~}\\ 
\end{tabular}}\vspace{-1mm}

\hspace{12mm}\subfloat[We present the distribution of the number of objects ($x$) in an image. Some images have no annotated objects, so the sum of percentages is not 100\%. \label{tab:results-coco-stats}]{
\tablestyle{2.5pt}{1.05}\begin{tabular}{l|ccc}
 Range & $0 < x \le 10$ & $10 < x \le 30$ & $30 < x \le 100$ \\
\shline
\#Images     & 3754    & 1124    & 74     \\
Percentage   & 75.08\% & 22.48\% & 1.48\% \\
\end{tabular}}\hspace{3mm}
\subfloat[Sharing weights among the Refiner modules introduces extra parameters and does not degrade the performance. For parameter efficiency, we share the parameters by default.\label{tab:results-ablation-sharing-weights}]{
\tablestyle{4.8pt}{1.05}\begin{tabular}{c|c|rrrr}
 Baseline & Share Weights? & AP & AP$_{75}$ & AP$_s$ & Params\\
\shline
DAB-DETR-R50 & \xmark       & 45.2 & 48.5 & 26.7 & 44.5 M \\
DAB-DETR-R50 & \cmark       & 45.2 & 48.6 & 26.6 & 44.2 M \\
\end{tabular}}\vspace{2mm}\\
\caption{We present ablation studies on the COCO val2017 split. We also show the distribution of the number of objects in an image.}
\label{tab:ablations}\vspace{-3mm}
\end{table*}

\subsection{Ablation Studies}
\label{subsec:ablation}

To investigate the behavior of the proposed RefineBox, we conducted several ablation studies on COCO. More results are available in the supplementary material.

\noindent \textbf{Training Cost.} During training, we freeze the well-trained object detectors and update only the parameters belonging to the refinement network. The frozen detection models require no time and memory to calculate and save the gradients. As shown in Table~\ref{tab:results-training-time}, it only requires 58\% of the training time of DAB-DETR-R50 if we generate features and prediction results from the detector online. These properties benefit researchers whose computing resources are limited.

\noindent \textbf{Inference Cost.} Our refinement network is efficient as it introduces only 0.5 M parameters and 6.5 GFLOPs~\footnote{For Swin-Tiny, the additional parameter and FLOPs are 0.4 M and 5.5 G because the channels of the output feature pyramid differ from the ResNet.} (Table~\ref{tab:results-overhead}). The extra parameters and FLOPs account for 1.1\% and 7.3\% of DAB-DETR, respectively. For ResNet-101 and Swin-Tiny, the numbers will be further decreased: for ResNet-101, the percentage of extra parameters and FLOPs reduced to 0.8\% and 4.2\%. 

\noindent \textbf{Training Epochs.} We investigate the training epochs of our RefineBox. We choose the commonly used settings: 12, 24, 36, and 50 epochs. Results listed in Table~\ref{tab:results-ablation-epochs} indicate a 12 epochs training schedule is sufficient to produce impressive performance. It is expected as we build our refinement network on top of a well-trained object detector. 

We also notice that longer training time gives limited gains compared with the 12 epochs training setting. For example, compared with the basic setting (12 training epochs), training for 24 and 50 epochs gives an additional 0.4 AP and 0.6 AP improvements, respectively. We suspect that this phenomenon should be attributed to the low capacity of the refinement network: the total number of parameters of the refinement network is 0.5 M, which is much less than the well-trained object detector.

\noindent \textbf{Refining Top $K$ Boxes.} We refine the top 100 predictions based on the classification logits. We also investigate the influence of $K$ and present results in Table~\ref{tab:results-ablation-topk}. We observe no gains when increasing the $K$ to 300. This phenomenon can be explained by the maximum number of objects that may exist in an image of the COCO dataset is 100. 

We then try to reduce the value of $K$ from 100 to 30. Interestingly, the loss of AP is small, \ie, from 45.2 AP to 45.0 AP. Refining only the top 30 predictions gives a 1.7 AP improvement on DAB-DETR-R50. This is because only 1.48\% of images contain more than 30 objects, as illustrated in Table~\ref{tab:results-coco-stats}. The FLOPs are also lowered by 21.5\%, resulting in 5.1 extra GFLOPs over DAB-DETR-R50. 

Our RefineBox suffers a 0.7 AP decrease if we set the $K$ as 10 but still outperforms DAB-DETR-R50 by 1.2 AP. We suspect that the top 10 prediction results contain false positives. In addition, the number of objects appearing in an image may exceed 10, though most images have no more than 10 objects, as shown in Table~\ref{tab:results-coco-stats}. Although smaller $K$ leads to lower FLOPs, we still set $K$ to the maximum number of objects that may appear in an image, which is 100 for COCO.

\noindent \textbf{Sharing Weights.} We conduct experiments on sharing weights between the Refiner modules. Results in Table~\ref{tab:results-ablation-sharing-weights} show that sharing parameters does not influence performance. As a result, we choose to share weights for parameter efficiency. 

\section{Conclusion}
\label{sec:conclusion}
We analyze various DETR-like models to seek methods to improve object detection performance by reducing the training errors of positive samples. We find that the localization ability is a bottleneck for current DETR-like detectors. Based on this observation, we propose a simple, flexible, and general framework called RefineBox to improve the localization capability of DETR-like models. We demonstrate the effectiveness and efficiency of our framework. In the future, we will concentrate on studying more effective methods to guide the refinement network to reduce localization errors. We plan to explore reinforcement learning with rewards as feedback for object detection, especially reinforcement learning from human feedback.

{\small
\bibliographystyle{ieee_fullname}
\bibliography{egbib}
}

\newpage
\appendix

\begin{figure}[!t]
    \subfloat[The ideal performance on AP$_{50}$.\label{fig:ideal-performance-ap50}]{
        \includegraphics[trim={0.3cm 0.0cm 0.0cm 0.0cm}, clip, width=0.98\linewidth]{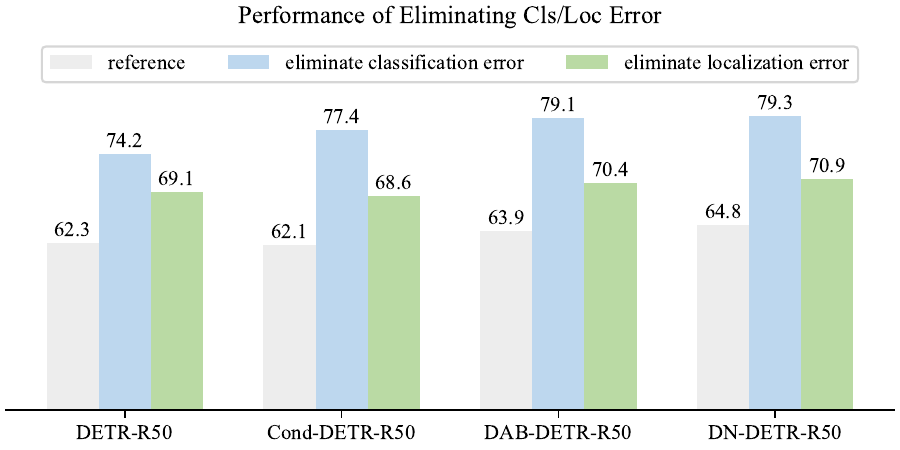}
    }\\
    \subfloat[The ideal performance on AP$_{75}$.\label{fig:ideal-performance-ap75}]{
        \includegraphics[trim={0.3cm 0.0cm 0.0cm 0.0cm}, clip, width=0.98\linewidth]{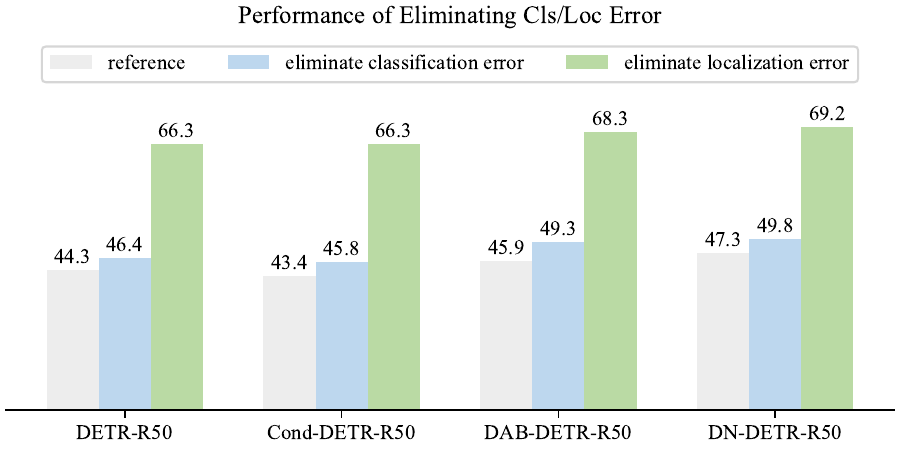}
    }\vspace{3mm}
    \caption{Compared to AP$_{75}$, the localization issue on AP$_{50}$ is relatively less severe. We suppose this is a potential reason why the performance improvement on AP$_{50}$ is less compared to AP$_{75}$.}
    \label{fig:ideal-performance}
\end{figure}

\section{About the Improvements on AP$_{50}$ and AP$_{75}$}
Compared with AP$_{75}$ and the main metric AP, the improvements on AP$_{50}$ are relatively lower. For instance, Conditional-DETR-R50~\cite{Conditional-DETR-ICCV-2021} on COCO~\cite{COCO-ECCV-2014} shows an increase of 2.5 on AP and 1.0 on AP${50}$. Recall how we investigate the bottleneck in the main paper (Section 3), we visualize the ideal performance on AP$_{50}$ and AP$_{75}$ without classification or regression errors in Figure~\ref{fig:ideal-performance}. The relatively less severe localization issue on AP$_{50}$ as illustrated in Figure~\ref{fig:ideal-performance-ap50} may be a contributing factor to the lower improvements on this metric.

The ideal performance on AP$_{75}$ in Figure~\ref{fig:ideal-performance-ap75} is consistent with our main observation in the paper: improving the localization performance results in greater gains than eliminating classification errors. Table~\ref{tab:gains-on-AP75} demonstrates a significant improvement in the ideal performance (without classification errors) by applying our RefineBox. This implies the proposed RefineBox relieves the localization problem when the required IoU is set as 0.75.

\begin{table}[!t]
\tablestyle{2pt}{1.1}
\begin{tabular}{l|cccc}
Model                      & Reference & Original & + RefineBox & $\Delta$ \\
\shline
DETR-R50~\cite{Deformable-DETR-arXiv-2020}             & 44.3 & 46.4 & \textbf{52.1} & \impro{+5.7} \\
Conditional-DETR-R50~\cite{Conditional-DETR-ICCV-2021} & 43.4 & 45.8 & \textbf{52.5} & \impro{+6.7} \\
DAB-DETR-R50~\cite{Deformable-DETR-arXiv-2020}         & 45.9 & 49.3 & \textbf{54.0} & \impro{+4.7} \\
DN-DETR-R50~\cite{DN-DETR-CVPR-2022}                   & 47.3 & 49.8 & \textbf{54.3} & \impro{+4.5} \\
\end{tabular}
\vspace{2mm}
\caption{Our RefineBox significantly improves the ideal performance (without classification errors). The actual performances of detectors are shown in the \textit{Reference} column. \textit{Original}: detectors without classification errors and RefineBox. \textit{+ RefineBox}: applying the RefineBox to the detection models.}
\label{tab:gains-on-AP75}
\vspace{-.5em}
\end{table}


\begin{figure}[!t]
\begin{center}
\includegraphics[trim={8.0cm 6.0cm 8.0cm 6.0cm}, clip, width=0.48\textwidth]{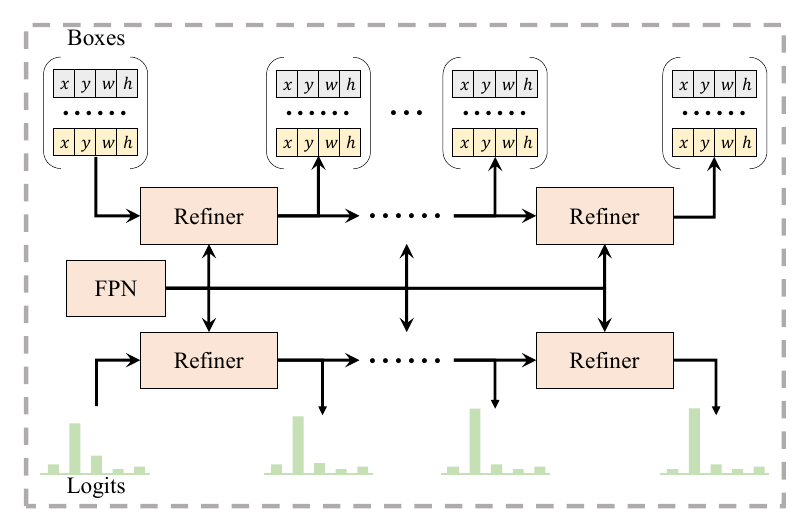}
\end{center}
\vspace{-1.0em}
   \caption{The refinement network with additional classification branch. The details of the Refiner modules are available in the paper.}
\label{fig:refiner-with-logits}
\end{figure}

\begin{table*}[!ht]\vspace{1mm}\hspace{0mm}
\subfloat[\textbf{Number of Refiner modules $M$.} More Refiner modules bring better performance. \label{tab:results-ablation-refiner}]{
\tablestyle{2.5pt}{1.05}\begin{tabular}{l|crr}
        Model & $M$ & AP & GFLOPs\\
        \shline
        Cond-DETR~\cite{Conditional-DETR-ICCV-2021} & - & 41.0 & 87.9 \\
        + RefineBox                                 & 1 & 43.2 & 93.0 \\
        + RefineBox                                 & \defaultchoice{3} & \defaultchoice{43.5} & 94.4 \\
        + RefineBox                                 & 6 & \textbf{43.6} & 96.4\\
         \multicolumn{4}{c}{~}\\
         \multicolumn{4}{c}{~}\\ 
\end{tabular}}\hspace{2mm}
\subfloat[\textbf{Batch Size.} Our RefineBox supports training with larger batch sizes, which leads to better performance. \label{tab:results-ablation-batch-size}]{
\tablestyle{4.8pt}{1.05}
    \begin{tabular}{l|cr}
        Model & batch size & AP \\
        \shline
        DETR-R50~\cite{DETR-ECCV-2020}    &  - & 42.0 \\
        + RefineBox & \defaultchoice{16} & \defaultchoice{44.4} \\
        + RefineBox & 64 & \textbf{44.7} \\
        \hline 
        DETR-R101~\cite{DETR-ECCV-2020}   &  - & 43.5 \\
        + RefineBox & \defaultchoice{16} & \defaultchoice{45.5} \\
        + RefineBox & 64 & \textbf{45.8} \\
\end{tabular}}\hspace{2mm}
\subfloat[\textbf{Bottleneck Channels.} Larger bottleneck channels bring greater gains but also entail higher costs.\label{tab:results-ablation-bottleneck-channels}]{
\tablestyle{4pt}{1.05}
    \begin{tabular}{l|c|rrr}
        Model          & \#channels & AP  & Params & GFLOPs \\
        \shline
        DAB-DETR-R50~\cite{DAB-DETR-arXiv-2022}                                &  -  & 42.0 & 43.7 M &  89.2 \\
        + RefineBox & 32  & 44.9 & 44.1 M &  94.3 \\
        + RefineBox & \defaultchoice{64}  & \defaultchoice{45.2} & \defaultchoice{44.2 M} &  \defaultchoice{95.7} \\
        + RefineBox & 128 & 45.4 & 44.6 M & 101.0 \\
        + RefineBox & 256 & \textbf{45.5} & 45.9 M & 121.3 \\
        \multicolumn{5}{c}{~}\\
\end{tabular}}\vspace{-1mm}

\hspace{13mm}\subfloat[\textbf{Number of Residual Blocks.} Considering trade-offs, we set the number of Residual blocks to 3 in each Residual Module. \label{tab:results-ablation-bottleneck-number}]{
\tablestyle{4pt}{1.05}
    \begin{tabular}{l|c|rrr}
         Model            & \#blocks & AP  & Params & GFLOPs \\
        \shline
        DAB-DETR-R50~\cite{DAB-DETR-arXiv-2022}                                      &  -  & 42.0 & 43.7 M &  89.2 \\
        + RefineBox & 1   & 45.0 & 44.1 M &  94.4 \\
        + RefineBox & 2   & 45.1 & 44.2 M &  95.0 \\
        + RefineBox & \defaultchoice{3}   & \defaultchoice{\textbf{45.2}} & \defaultchoice{44.2 M} &  \defaultchoice{95.7} \\
        + RefineBox & 6   & 45.1 & 44.3 M &  97.7 \\
\end{tabular}}\hspace{8mm}
\subfloat[\textbf{Model Dimension.} Larger model dimension brings more improvements to the baseline, but it also leads to a greater computational burden.\label{tab:results-ablation-model-dimension}]{
\tablestyle{4pt}{1.05}
    \begin{tabular}{l|c|rrr}
        Model                   & dim & AP  & Params & GFLOPs \\
        \shline
        DAB-DETR-R50~\cite{DAB-DETR-arXiv-2022}                                         &  -     & 42.0 & 43.7 M &  89.2 \\
        + RefineBox          & 32     & 44.6 & 43.9 M &  91.3 \\
        + RefineBox          & \defaultchoice{64}     & \defaultchoice{45.2} & \defaultchoice{44.2 M} &  \defaultchoice{95.7} \\
        + RefineBox          & 128    & 45.3 & 45.3 M & 111.6 \\
        + RefineBox          & 256    & \textbf{45.5} & 49.3 M & 171.7 \\
\end{tabular}}\vspace{2mm}
\caption{We conduct more ablation studies to investigate the influence of different factors. Our default choices are in gray.}
\label{tab:results-ablations}\vspace{-3mm}
\end{table*}

\section{RefineBox with Classification Branch}
In Section 5.3 of our main paper, we delve deeper into \textit{whether an additional classification refinement branch is beneficial}. As shown in Figure~\ref{fig:refiner-with-logits}, the classification refinement branch is structured similarly to the localization refinement branch, with separate parameters and Refiner modules. Additional information about the Refiner module can be found in the main paper's Figure 4.
\section{More Ablation Studies}

\noindent \textbf{Number of Refiner Modules.} Table~\ref{tab:results-ablation-refiner} showcases the ablation studies on the number of Refiner modules used in the RefineBox. The results indicate that our RefineBox performs well across various Refiner module counts, including 1, 3, and 6. As we observe the performance gains tend to saturate, we choose 3 as the value of the number of Refiner modules.

\noindent \textbf{Large Batch Size.} Our proposed RefineBox offers the advantage of enabling training with larger batch sizes without the need for memory optimization, which can be especially beneficial for researchers with limited resources. For instance, when employing the detector as an online feature extractor and region proposal network, our RefineBox allows training DAB-DETR-R50 + RefineBox with a total batch size of 160 on 4 GPUs, each equipped with 24GB of memory.

We conducted experiments on larger batch sizes with DETR, as presented in Table~\ref{tab:results-ablation-batch-size}. The results show that using a batch size of 64 yields slightly better performance than a batch size of 16. For both DETR-R50 and DETR-R101, increasing the batch size from 16 to 64 leads to a 0.3 increase in AP.

\noindent \textbf{Bottleneck Channels.} As mentioned in the paper, we denote the input channels of the $3 \times 3$ conv layer in the Bottleneck Block as the bottleneck channels. We perform ablation studies on the bottleneck channels to strike a balance between performance and inference cost. From the results in Table~\ref{tab:results-ablation-bottleneck-channels}, we observe that increasing the bottleneck channels results in improved AP performance. However, compared to the default value of 64, increasing the bottleneck channels to 128 only provides a modest gain of 0.2 AP while introducing 5.3 GFLOPs, which does not seem like an optimal trade-off. Therefore, we choose 64 as the default bottleneck channel size.

\noindent \textbf{The Number of Bottleneck Blocks.} Table~\ref{tab:results-ablation-bottleneck-number} presents the ablation study on the number of bottleneck blocks in our RefineBox. The results show that the performance of RefineBox with different numbers of bottleneck blocks is comparable, indicating that adding more blocks may not significantly contribute to the learning of features related to localization. These findings suggest that there is room for further exploration in designing the bounding box refinement module of our RefineBox framework. We leave the investigation of alternative architectures for future work.

\noindent \textbf{Model Dimension.} We ablate the model dimension in Table~\ref{tab:results-ablation-model-dimension}. As we have introduced in the paper, the model dimension is the input and output channels of modules in the refinement module, except for FPN. This value is the key to restricting the GFLOPs and the number of parameters of the refinement module. Setting the dimension as 32 leads to the lightest version of RefineBox, but the AP is also the worst. The converse approach is to increase the value of the model dimension, which leads to only marginal performance gains but introduces a significant number of FLOPs. Considering the trade-off between performance and computational cost, we chose 64 as the default model dimension.

\begin{table}
\tablestyle{2pt}{1.1}
\begin{tabular}{l|rrrr}
Model                      & AP & AP$_{75}$ & AP$_{s}$ & AR$_{s}$ \\
\shline
Deforamble-DETR~\cite{Deformable-DETR-arXiv-2020}                                                         
                           & 44.7 & 48.8 & 27.0 & 46.0 \\
+ RefinerBox               & 45.3 & 49.5 & 27.8 & 48.0 \\
$\Delta$                   & +0.6 & +0.7 & +0.8 & +2.0 \\
\hline
Deformable-DETR-Two-Stage~\cite{Deformable-DETR-arXiv-2020}                                               
                           & 47.1 & 51.3 & 30.9 & 52.0 \\
+ RefineBox                & 47.3 & 51.5 & 31.2 & 52.8 \\
$\Delta$                   & +0.2 & +0.2 & +0.3 & +0.8 \\
\end{tabular}
\vspace{2mm}
\caption{The simple refinement network in our experiments brings little gains on the \textit{models which have well utilized the multi-scale features.}}
\label{tab:multi-scale-features}
\vspace{-.5em}
\end{table}
\section{More Discussion on Multi-scale Features}
We have hypothesized that the refinement network in our experiments improves the DETR-like models by efficiently and effectively leveraging the multi-scale backbone features, and more sophisticated designs may bring more gains. Here we present more discussion on the utilization of the multi-scale features.

It's interesting to see that the gains from the refinement network are higher on single-scale models compared to those that utilize multi-scale features. For example, as shown in Table~\ref{tab:multi-scale-features}, the refinement network in our experiments only gives 0.6 AP improvement for Deformable DETR~\cite{Deformable-DETR-arXiv-2020}. We notice that each Transformer layer in the Deformable DETR utilizes the multi-scale backbone features with the help of the sophisticated Multi-scale Deformable Attention mechanism. It combines the best of the sparse spatial sampling of deformable convolution and the relation modeling capability of Transformers and enables feature aggregation across the inter-scale and intra-scale features. We refer to Deformable DETR~\cite{Deformable-DETR-arXiv-2020} for more details of the Deformable Attention mechanism.

In contrast, the refinement network in our experiments is quite simple, which only adopts a sequence of residual blocks to refine the boxes. We suspect the model capability of the refinement network may not be much better than the Multi-scale Deformable Attention mechanism. Given the simplicity of our refinement network, it's not a surprise that our model gives limited improvements on such multi-scale models. 

Please note that the goal of our RefineBox is to explore the potential of refining the boxes of well-trained DETR-like models instead of exploring a better multi-scale features aggregation mechanism. We hope that our work can inspire future research in this area and lead to more powerful refinement networks and object detection models.


\end{document}